\documentclass[10pt,journal,compsoc]{IEEEtran}

\usepackage[T1]{fontenc}
\usepackage{newtxtext,newtxmath}

\usepackage{amsmath,amssymb}
\usepackage{booktabs}
\usepackage{graphicx}
\usepackage{algorithm}
\usepackage{algpseudocode}
\usepackage{tikz}
\usetikzlibrary{arrows.meta,positioning,fit}
\usepackage{pgfplots}
\usepgfplotslibrary{groupplots}
\pgfplotsset{compat=1.18}
\usepackage[hidelinks]{hyperref}
\usepackage{url}
\usepackage{balance}

\newcommand{\method}{LABELSENSE-Pilot}

\newcommand{\one}{\mathbf{1}}
\newcommand{\SourceRecords}{2,500}

\newcommand{\LSDisplayPct}{85.62}
\newcommand{\LSFlickerPct}{2.09}
\newcommand{\LSPrefPct}{76.47}
\newcommand{\LSRuntimeMs}{27.01}
\newcommand{\ILPDisplayPct}{87.05}
\newcommand{\ILPFlickerPct}{14.13}

\newcommand{\FlickerReductionVsILP}{12.04}
\newcommand{\FlickerRelativeReductionVsILP}{85.2}
\newcommand{\PreferenceGainVsILP}{2.77}
\newcommand{\DisplayGainVsGraph}{17.24}
\newcommand{\RuntimeRatioVsGraph}{61.4}
\newcommand{\DynFlickerPct}{0.48}
\newcommand{\AwareFifteenDisplayPct}{79.50}
\newcommand{\AwareFifteenViolationPct}{0.00}
\newcommand{\UnawareViolationPct}{52.57}
\newcommand{\UnawareAblationViolationPct}{66.35}
\newcommand{\DensityEightyDisplayPct}{56.25}
\newcommand{\DensityTwentyRuntimeMs}{7.39}
\newcommand{\DensityEightyRuntimeMs}{305.96}
\newcommand{\DensityEightyRuntimeSdMs}{163.35}
\newcommand{\FullAblationFlickerPct}{1.61}
\newcommand{\NoTemporalFlickerPct}{12.93}
\newcommand{\TemporalFlickerReduction}{11.31}

\title{Constraint-Safe Graph-Context Scoring for Stable Point-Feature Labels Under Text-Width and Accessibility-Inspired Profiles}

\author{Taimoor Ahmed \thanks{Superior University Lahore, Pakistan}}
\begin{document}
\maketitle

\begin{abstract}
Point-feature label placement on interactive maps must reconcile geometric validity, display yield, local placement utility, and stability across camera motion. Accessibility and multilingual requirements further change label dimensions, yet algorithmic evaluations often collapse these concerns into overlap counts. We present LABELSENSE-Pilot, a reproducible prototype that generates eight compass candidates per feature, scores candidates with a multilayer perceptron over graph-context summaries, adds a previous-placement bonus, and selects a layout through mixed-integer optimization. The executed scorer is deliberately not described as a graph transformer. Every returned layout is checked for viewport containment, per-feature uniqueness, and pairwise clearance. Experiments use 2,500 airport coordinates and names spanning 155 countries, with country-grouped splits and generated density, camera, text-suffix, preference, and enlarged-font stressors. Across five seeds, LABELSENSE-Pilot displayed 85.62 percent of labels with 2.09 percent flicker and zero collisions. Versus a handcrafted-utility integer program, LABELSENSE-Pilot sacrificed 1.43 percentage points of display while reducing flicker by 12.04 points. Enlarged-box-aware layouts produced zero proxy violations, whereas standard geometry reevaluated at 1.5x violated 52.57 percent of selected placements. These results establish an auditable engineering trade-off, not human accessibility, multilingual usability, or preference. Official recent baselines and participant evidence remain required before submission.

\end{abstract}


\section{Introduction}
\label{sec:introduction}
Placing text near point features is deceptively difficult. A label should remain inside the viewport, avoid every other selected label, stay close to its anchor, respect feature priority, and use a position that readers can associate with the point. On an interactive map, the solution also has memory: a layout that is geometrically valid in every frame may still be difficult to follow when positions switch after small pans or zooms. Label width changes under localization and font enlargement make the feasible set itself dependent on the rendering profile. These coupled effects motivate a placement system that separates inviolable geometry from configurable utility.

Point-feature labeling has long been formulated through finite candidate positions, conflict graphs, heuristics, and combinatorial optimization \cite{christensen1995empirical,zoraster1990solution,klau2003optimal,a6,a5,a8,haunert2017beyond}. Recent work adds reinforcement learning for map labels \cite{bobak2024reinforced}, graph-based learning in the adjacent illustration-labeling domain \cite{qu2025graph}, explicit preference evidence \cite{scheuerman2023visual}, and temporal consistency \cite{gemsa2020unified,a9,a10,bonerath2025timeslider}. Accessibility research also makes clear that readable web maps require more than collision avoidance \cite{manu2025accessibility,a11,a12,hennig2017accessible}. A single system that claims to unify all these strands therefore carries an unusually high evidence burden. In particular, a geometry check cannot stand in for a study with low-vision readers, and a few script-bearing strings cannot establish multilingual usability.

This paper reports a deliberately bounded algorithmic pilot. \method{} generates eight axis-aligned candidate rectangles, represents local competition with three fixed graph-context summaries, learns a candidate utility through a small multilayer perceptron (MLP), applies a previous-direction bonus, and uses a binary mixed-integer program (MILP) to enforce viewport and pairwise-clearance constraints. The implementation is not an end-to-end graph transformer. Its preference targets, camera paths, density patterns, generic facility suffixes, and enlarged-font profiles are generated stressors. Real evidence is limited to the coordinates and source names of a pinned 2,500-airport fixture. These boundaries are part of the result rather than fine print.

The pilot asks three engineering questions. First, can learned graph-context scoring plus exact feasibility display more labels than its learned greedy decoder? Second, does a temporal bonus reduce direction changes relative to independent-frame scoring? Third, does constructing candidates at the evaluation font scale prevent enlarged-box violations that an unaware layout incurs? The evaluation also records where the method loses: a handcrafted MILP displays slightly more labels, dynamic greedy is more stable, and both recent-inspired greedy proxies are far faster.

The contributions are:
\begin{itemize}
\item An auditable hybrid architecture that combines fixed graph-context features and an explicitly identified MLP surrogate with a hard-constrained MILP decoder and an independent geometric audit.
\item A reproducible five-seed pilot with country-grouped data splits, matched candidates, six local comparators, ten quantitative panels, synchronized result tables, and declared synthetic stressors for time, text width, preference, and enlarged geometry.
\item A claim-and-submission gate that distinguishes collision-free computation from human-centered evidence and records the unexecuted graph-transformer, official-baseline, urban-data, authentic-translation, and participant-study requirements.
\end{itemize}

The remainder of this paper is organized as follows. Section~\ref{sec:related} positions the pilot against candidate-based, learned, dynamic, preference, and accessibility research. Section~\ref{sec:method} defines the executed scoring and constrained decoder. Section~\ref{sec:experiments} reports the protocol and results. Section~\ref{sec:discussion} interprets the trade-offs, threats, and submission gate, and Section~\ref{sec:conclusion} concludes.

\section{Related Work}
\label{sec:related}

\subsection{Candidate-Based and Learned Placement}

The classical point-feature problem selects at most one candidate from a finite set for each point. Its conflict graph exposes incompatible rectangles, while position, anchor distance, and feature priority define soft value. Early empirical work compared greedy, gradient-descent, and simulated-annealing strategies \cite{christensen1995empirical}. Exact formulations include large 0--1 cartographic programs, optimal rectangular-label models, and extended integer programs beyond a plain maximum independent set \cite{zoraster1990solution,klau2003optimal,haunert2017beyond}. Recent parallel and spatial-pattern methods broaden the scale--quality trade space through map segmentation, fixed--sliding hybrids, and data-mined placement order \cite{lessani2021parallel,lessani2025mpi,cao2023spatial}; GPU-parallel greedy evaluation targets interactive throughput \cite{pavlovec2022rapid}. Our decoder remains in this established finite-candidate family. ``Exact'' refers to the encoded constraints of a returned, independently checked incumbent, not polynomial runtime or a universal cartographic optimum.

Learned systems can model interactions omitted by handcrafted scores. Reinforced Labels applies multi-agent deep reinforcement learning to point-feature placement \cite{bobak2024reinforced}. LPGT and its retrieval-augmented LPCE successor learn interactions for appliance-manual illustrations, not maps \cite{qu2025graph,zhang2026lpce}. RL-LABEL addresses dynamic augmented-reality labels \cite{chen2023rllabel}. These adjacent architectures motivate context-sensitive scoring, but the executed model here is materially smaller: three deterministic neighborhood summaries feed a conventional two-hidden-layer MLP. It neither reproduces those systems nor substantiates a transformer claim.

\subsection{Temporal Consistency}

Dynamic-map research distinguishes per-frame validity from sequence quality. Foundational work formulated consistent pan/zoom labeling and optimized active ranges \cite{been2006dynamic,been2010active}. A later experimental framework and its journal development formalized temporal intervals and algorithms \cite{barth2016temporal,gemsa2020unified}; they form a related version family rather than two independent novelty signals. Hybrid optimization addresses smoothness \cite{he2022smoothness}, transition research studies changing point-label geometry \cite{depian2023transitions}, and fully dynamic independent-set work handles evolving conflict graphs \cite{bhore2022dynamic}. Recent methods target fixed-scale dense exploration, zoom-aware tabu search, and time-slider consistency \cite{gedicke2021zoomless,cutello2025tabu,bonerath2025timeslider}. \method{} uses only the previous compass direction. It therefore tests short-horizon switching on scripted motion, not global temporal consistency, transitions, or observed interaction behavior.

\subsection{Preference, Language, and Accessibility}

Placement preference is not interchangeable with a conventional top-right rule. Scheuerman \emph{et al.} published explicit binary and ranking judgments plus a companion data record \cite{scheuerman2023visual,scheuerman2023dataset}. The current execution did not acquire those responses and instead uses three disclosed synthetic rankings. Production audits show that authentic label geometry and association remain practical cartographic challenges \cite{harrie2022challenges}. Our five suffix conditions exercise only a character-class width heuristic; they do not implement browser shaping, validated translation, bidirectionality, or font fallback.

Accessibility work covers typography, contrast, interaction, adaptation, and nonvisual exploration. Large-print SVG maps and general accessible-web-map recommendations predate this pilot \cite{kulyukin2010svg,hennig2017accessible}; AltGeoViz supports accessible geovisualization \cite{li2024altgeoviz}; recent indoor-map adaptations and thematic-web-map criteria add disabled-community and low-vision perspectives \cite{madugalla2025adaptive,manu2025accessibility}. Enlarging rectangles and auditing overlap is consequently only an accessibility-inspired geometry proxy. It does not measure readability, comprehension, error, or task performance.

Table~\ref{tab:closest-work} summarizes the nearest capabilities conservatively. A check mark denotes evidence in the cited setting, not matched semantics or an official reproduction. The final row records what this artifact actually executes.

\begin{table*}[t]
\caption{Closest-work capability matrix. ``Hard'' denotes explicit feasibility constraints; ``human'' denotes participant-derived evidence. Proxy outcomes do not qualify as human evidence.}
\label{tab:closest-work}
\centering
\scriptsize
\setlength{\tabcolsep}{3.5pt}
\begin{tabular}{lcccccc}
\toprule
Work & Learned & Dynamic & Preference & Text/profile & Hard & Human \\
\midrule
Christensen \emph{et al.} \cite{christensen1995empirical} & -- & -- & rule & -- & -- & -- \\
Zoraster; Klau--Mutzel; Haunert--Wolff \cite{zoraster1990solution,klau2003optimal,haunert2017beyond} & -- & -- & cost & -- & \checkmark & -- \\
Lessani \emph{et al.}; Cao \emph{et al.} \cite{lessani2021parallel,lessani2025mpi,cao2023spatial} & -- & -- & cost & -- & repair & -- \\
Rapid Labels \cite{pavlovec2022rapid} & -- & -- & cost & -- & greedy & -- \\
Reinforced Labels \cite{bobak2024reinforced} & \checkmark & -- & learned & geometry & learned & -- \\
LPGT / LPCE \cite{qu2025graph,zhang2026lpce} & \checkmark & -- & learned & geometry & learned & -- \\
RL-LABEL \cite{chen2023rllabel} & \checkmark & \checkmark & learned & AR & learned & \checkmark \\
Been \emph{et al.} \cite{been2006dynamic,been2010active} & -- & \checkmark & cost & -- & model & -- \\
Barth / Gemsa \emph{et al.} \cite{barth2016temporal,gemsa2020unified} & -- & \checkmark & cost & -- & model & -- \\
He; Depian \emph{et al.} \cite{he2022smoothness,depian2023transitions} & -- & \checkmark & cost & -- & mixed & -- \\
Recent dynamic algorithms \cite{bhore2022dynamic,gedicke2021zoomless,cutello2025tabu,bonerath2025timeslider} & -- & \checkmark & cost & -- & mixed & -- \\
Scheuerman \emph{et al.} \cite{scheuerman2023visual,scheuerman2023dataset} & -- & -- & \checkmark & -- & -- & \checkmark \\
Accessibility/geovisualization \cite{harrie2022challenges,kulyukin2010svg,hennig2017accessible,li2024altgeoviz,madugalla2025adaptive,manu2025accessibility} & -- & -- & mixed & \checkmark & -- & \checkmark \\
\midrule
\method{} & MLP surrogate & 1-frame & synthetic & proxy & \checkmark & -- \\
\bottomrule
\end{tabular}
\end{table*}

The defensible gap is narrower than the aspirational concept. This pilot contributes a transparent integration of learned local utility, one-step persistence, profile-dependent rectangles, and constraint-safe decoding under a reproducible protocol. It does not establish the novelty or benefit of a graph transformer, real preference learning, authentic multilingual layout, or accessible map use; those remain conditional extensions requiring direct evidence.

\section{Method}
\label{sec:method}

\subsection{Problem and Notation}

At frame $t$, feature $i\in\mathcal I_t$ has screen anchor $a_{it}=(x_{it},y_{it})$, source name $s_i$, priority $p_i\in[0,1]$, and optionally a compass direction selected at $t-1$. The viewport is $V=[0,W]\times[0,H]$ with $W=1024$ and $H=768$ in the experiments. Each point receives the eight compass candidates $\mathcal K=\{N,NE,E,SE,S,SW,W,NW\}$ plus implicit non-display. Table~\ref{tab:notation} gives the principal symbols.

\begin{table}[t]
\caption{Notation for the executed pilot.}
\label{tab:notation}
\centering
\scriptsize
\begin{tabular}{ll}
\toprule
Symbol & Meaning \\
\midrule
$a_{it},s_i,p_i$ & screen anchor, source name, priority \\
$g,\alpha$ & suffix condition, font-scale profile \\
$R_{ikt}$ & candidate rectangle \\
$F_{ikt}$ & viewport-feasibility indicator \\
$h_{ikt}$ & 12-dimensional candidate feature vector \\
$\widehat u_{ikt}$ & MLP-predicted synthetic utility \\
$z_{ikt}$ & previous-direction indicator \\
$x_{ikt}$ & binary candidate-selection variable \\
$E_t$ & inter-feature rectangle conflicts \\
$\delta$ & required box clearance in pixels \\
\bottomrule
\end{tabular}
\end{table}

The suffix condition $g$ appends a controlled facility term while preserving the proper-name stem. A character-class function $\omega(c)$ assigns 0.33 units to spaces, 1.00 to CJK characters, 0.68 to Arabic characters, 0.85 to selected wide Latin symbols, 0.32 to narrow symbols, and 0.58 otherwise. With font size $f$ and enlargement factor $\alpha$, the heuristic dimensions are
\begin{equation}
\begin{aligned}
q(s_i,g)&=\max\!\left(1,\sum_{c\in s_i\oplus g}\omega(c)\right),\\
(w_i,h_i)&=(q\alpha f+8,1.35\alpha f+4).
\end{aligned}
\label{eq:size}
\end{equation}
This is deterministic width stress; it is not glyph shaping, translation validation, or a perceptual accessibility model.

For direction $k$, let $(d_x^k,d_y^k)\in\{-1,0,1\}^2$ be its unit compass offset and $o=4$ pixels. The candidate center and rectangle are
\begin{equation}
\begin{aligned}
c_{ikt}&=a_{it}+\left(d_x^k(w_i/2+o),d_y^k(h_i/2+o)\right),\\
R_{ikt}&=c_{ikt}+[-w_i/2,w_i/2]\times[-h_i/2,h_i/2].
\end{aligned}
\label{eq:rect}
\end{equation}
Viewport validity is precomputed as
\begin{equation}
F_{ikt}=\one\{0\le R^x_{0},\;R^x_{1}\le W,\;0\le R^y_{0},\;R^y_{1}\le H\}.
\label{eq:viewport}
\end{equation}

\subsection{Preference and Graph-Context Features}

The three profiles (balanced, right, and upper) define complete, disclosed rankings of the eight directions. If $r_\pi(k)\in\{0,\ldots,7\}$ is the rank under profile $\pi$, the proxy agreement feature is
\begin{equation}
P_\pi(k)=1-\frac{r_\pi(k)}{7}.
\label{eq:preference}
\end{equation}
It is generated and should not be confused with the participant responses in \cite{scheuerman2023visual,scheuerman2023dataset}.

Graph context is computed over point anchors before candidate selection. For every other point $j$, let $d_{ij}=\lVert a_{it}-a_{jt}\rVert_2$. The three fixed summaries are
\begin{align}
L_i&=\frac{\sum_{j\ne i}e^{-d_{ij}/100}\,\bar w_j}
{\sum_{j\ne i}e^{-d_{ij}/100}},\nonumber\\
Q_i&=\frac{\sum_{j\ne i}e^{-d_{ij}/240}(0.5+p_j)p_j}
{\sum_{j\ne i}e^{-d_{ij}/240}(0.5+p_j)},\nonumber\\
M_i&=\frac{1}{|\mathcal I_t|-1}\sum_{j\ne i}e^{-d_{ij}/90},
\label{eq:context}
\end{align}
where $\bar w_j=\min(w_j/250,2)$. These exponential averages are fixed attention-like summaries; their scales are not trained. Candidate features concatenate bias, priority, preference, normalized anchor distance, viewport-edge clearance, 150-pixel density, normalized width, scale, previous-direction match, and $(L_i,Q_i,M_i)$:
\begin{equation}
\begin{aligned}
h_{ikt}=[&1,p_i,P_\pi(k),\bar d_{ikt},e_{ikt},\rho_{it},\\
&\bar w_i,\alpha,z_{ikt},L_i,Q_i,M_i]^{\mathsf T}.
\end{aligned}
\label{eq:features}
\end{equation}

There is no human-labeled utility target. Training instead uses a declared seeded proxy that combines the same interpretable quantities with Gaussian noise $\epsilon\sim\mathcal N(0,0.12^2)$:
\begin{align}
y_{ikt}^{\rm syn}={}&2.2p_i+2.4P_\pi(k)-0.75\bar d_{ikt}+0.50e_{ikt}\nonumber\\
&-1.25\rho_{it}-0.70L_i+0.35Q_i-1.10M_i\nonumber\\
&+0.55p_iP_\pi(k)-0.45\bar w_iM_i+\epsilon.
\label{eq:target}
\end{align}
Train countries produce eight scenes of 30, 35, or 40 points per seed under rotating suffix and profile settings, totaling 2,200 candidate examples. Features are standardized using training means $\mu$ and scales $\sigma$. The executed \texttt{GraphContextMLP} is a $12$--$20$--$10$--$1$ regressor with $\tanh$ hidden activations, Adam optimization, $\ell_2$ coefficient 0.02, early stopping, and a seeded 15\% internal validation fraction:
\begin{align}
\widetilde h_{ikt}&=(h_{ikt}-\mu)\oslash\sigma,\nonumber\\
r^{(1)}_{ikt}&=\tanh(b_1+W_1\widetilde h_{ikt}),\nonumber\\
r^{(2)}_{ikt}&=\tanh(b_2+W_2r^{(1)}_{ikt}),\nonumber\\
\widehat u_{ikt}&=b_3+W_3r^{(2)}_{ikt}.
\label{eq:mlp}
\end{align}

\begin{figure*}[t]
\centering
\includegraphics[width=\textwidth]{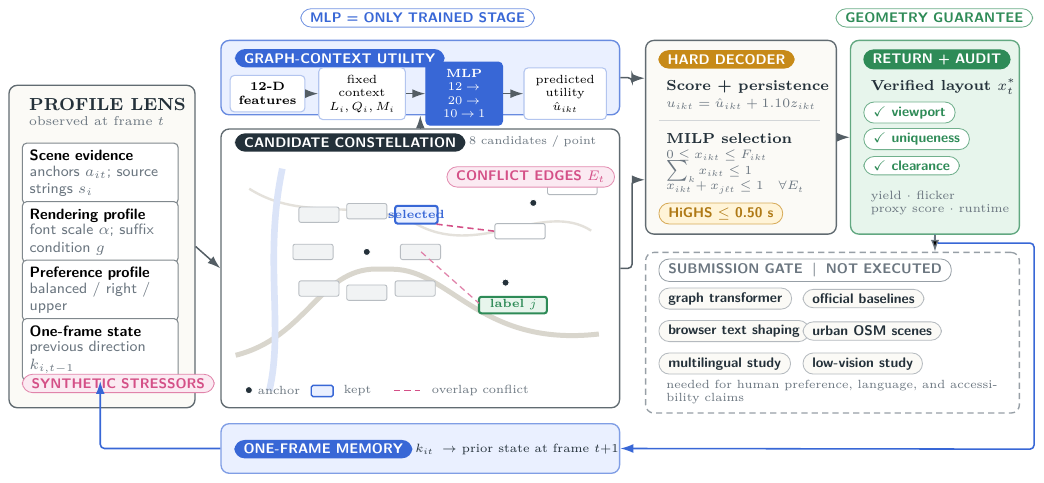}
\caption{Executed frame pipeline. Blue marks the only stage containing learned weights (the MLP); its graph summaries, box geometry, MILP constraints, and audit are deterministic.}
\label{fig:architecture}
\end{figure*}

\subsection{Temporal Score and Constraint-Safe Decoder}

The one-step state indicator equals one if object $i$ retains direction $k$ from the preceding frame. The final candidate utility is
\begin{equation}
\begin{aligned}
u_{ikt}&=\widehat u_{ikt}+\lambda_T z_{ikt},\\
z_{ikt}&=\one\{k=k_{i,t-1}\},\qquad \lambda_T=1.10.
\end{aligned}
\label{eq:temporal}
\end{equation}
The indicator rewards direction persistence; it does not penalize screen displacement after the known camera transform and it has no memory beyond one frame.

For each candidate, invalid viewport geometry is disabled by
\begin{equation}
0\le x_{ikt}\le F_{ikt},\qquad x_{ikt}\in\{0,1\}.
\label{eq:bounds}
\end{equation}
At most one position may be chosen for each feature:
\begin{equation}
\sum_{k\in\mathcal K}x_{ikt}\le 1\qquad \forall i\in\mathcal I_t.
\label{eq:one-position}
\end{equation}
Let $E_t$ contain candidate pairs from distinct features whose rectangles overlap after clearance $\delta=4$ pixels. Each conflict creates
\begin{equation}
x_{ikt}+x_{j\ell t}\le 1\qquad \forall\big((i,k),(j,\ell)\big)\in E_t.
\label{eq:conflicts}
\end{equation}
The decoder solves the binary program
\begin{align}
\max_{x}\quad &\sum_{i\in\mathcal I_t}\sum_{k\in\mathcal K}u_{ikt}x_{ikt}\nonumber\\
\text{s.t.}\quad&\text{Eqs.~\eqref{eq:bounds}--\eqref{eq:conflicts}}.
\label{eq:milp}
\end{align}
SciPy's HiGHS interface receives a 0.50-s time limit and zero requested relative gap. If no incumbent exists, the implementation invokes a feasible greedy fallback and records that status. Every selected incumbent is then audited independently. All main and reported sweep MILPs in this execution terminated optimally; the method nevertheless makes no general polynomial-time or global-runtime claim.

\subsection{Complexity and Guarantee Scope}

For $n$ visible features and eight candidates each, feature construction computes an $n\times n$ anchor-distance matrix and therefore costs $O(n^2)$ time and memory in this pilot. Pairwise rectangle testing is $O((8n)^2)$. MILP solution is NP-hard in the worst case. These choices favor clarity and auditability at the evaluated scale rather than asymptotic novelty. The returned layout has a direct, machine-checked guarantee only for the encoded axis-aligned rectangles, viewport, and clearance. It does not guarantee perceptual association, readable contrast, language correctness, or accessibility.
\begin{algorithm}[t]
\caption{Train graph-context utility surrogate}
\label{alg:training}
\begin{algorithmic}[1]
\Require country-grouped training records, seed $s$
\State initialize seeded generator; set suffixes and profiles
\For{each of eight training scenes}
  \State generate eight candidates and $h_{ikt}$ per point
  \State compute declared target $y_{ikt}^{\rm syn}$ by Eq.~\eqref{eq:target}
\EndFor
\State fit the standardizer on training candidates
\State fit seeded MLP by Eq.~\eqref{eq:mlp} with early stopping
\State \Return frozen standardizer, MLP, training RMSE, metadata
\end{algorithmic}
\end{algorithm}

\begin{algorithm}[t]
\caption{Place and audit one frame}
\label{alg:inference}
\begin{algorithmic}[1]
\Require anchors, strings, profile, previous directions
\State form rectangles by Eqs.~\eqref{eq:size}--\eqref{eq:viewport}
\State compute graph summaries and candidate features
\State predict utility and apply Eq.~\eqref{eq:temporal}
\State enumerate inter-feature conflict set $E_t$
\State solve Eq.~\eqref{eq:milp} within the recorded limit
\If{no MILP incumbent exists}
  \State construct a feasible utility-sorted greedy layout
\EndIf
\State recompute viewport, uniqueness, and clearance checks
\State \Return selected directions, timings, status, audit metrics
\end{algorithmic}
\end{algorithm}

\section{Experimental Evaluation}
\label{sec:experiments}

\subsection{Questions and Protocol}

The executed study tests three prespecified engineering hypotheses: \method{} should display more labels than the learned greedy decoder while preserving the encoded constraints (H1); the temporal term should reduce switching versus its removal (H2); and geometry-aware optimization should avoid enlarged-box violations that appear when 1.0$\times$ layouts are evaluated at 1.5$\times$ geometry (H3). None concerns human task performance.

\begin{table}[t]
\caption{Executed point-feature fixture and split. Coordinates and source names are real; camera motion, language suffixes, preference targets, and accessibility profiles are synthetic stressors.}
\label{tab:dataset}
\centering\small
\begin{tabular}{lr}
\toprule
Item & Count \\
\midrule
Point features & 2,500 \\
Countries & 155 \\
Train points (country-grouped) & 1,606 \\
Validation points & 292 \\
Test points & 602 \\
International-name heuristic & 368 \\
Main paired seeds & 5 \\
\bottomrule
\end{tabular}
\end{table}

The bundled fixture contains \SourceRecords{} unique airport records from 155 countries. Coordinates and source names are real; source rows identify OurAirports and are distributed in an OpenFlights-format file. The artifact records both providers' source pages and licensing notes. A deterministic country hash assigns 1,606 points to training, 292 to validation, and 602 to test, preventing points from one country appearing across splits. The main evaluation uses scenes from Brazil, Finland, Iran, Mexico, and Sweden. Airport points are not a substitute for dense urban OSM features; density sweeps replicate and jitter held-out points after splitting.

Five seeds (7, 19, 31, 43, 61) define one 42-point, five-frame sequence each. The main comparison therefore contains 30 method--seed sequences and 150 frames. All methods receive identical anchors, eight candidates, 12-pixel base font, 1.25$\times$ geometry, profile, and four-pixel clearance. The MLP is trained anew for each seed using training countries only. Its 2,200 synthetic targets per seed yield RMSEs from 0.120 to 0.124; this measures fit to the constructed target, not preference validity.

We compare Onion-Greedy; feasible 700-iteration simulated annealing (SimAnneal); an exact decoder over handcrafted utility (ILP-Utility); learned-utility greedy selection (GraphGreedy-Reimpl); learned greedy selection with previous-direction hysteresis (DynHysteresis-Reimpl); and \method{}. The two names ending in Reimpl are project-local, recent-inspired proxies. An official recent implementation was not executed. They are not LPGT, a faithful dynamic-labeling reproduction, or evidence of parity with the corresponding publications. This missing comparison is a submission blocker.

Metrics are displayed labels divided by input points, pairwise collision count, mean anchor distance, synthetic preference agreement, frame-to-frame direction changes among labels visible in both frames (flicker), enlarged-box violations, and single-process wall time. An initial frame has zero defined switching by construction. Runtime includes candidate scoring and selection as recorded by the local method call but excludes data preparation and rendering. All quantitative claims below are generated from CSV outputs rather than transcribed manually.

\subsection{Main Trade-Off}

\begin{table*}[t]
\caption{Main five-seed pilot comparison. Display, flicker, and preference are percentages; runtime is milliseconds per frame. All methods produced zero counted geometric collisions under the supplied boxes. Bold and underline denote best and second best only within this local execution.}
\label{tab:main-results}
\centering\small
\begin{tabular}{lrrrrl}
\toprule
Method & Display $\uparrow$ & Flicker $\downarrow$ & Proxy pref. $\uparrow$ & Runtime $\downarrow$ & Evidence status \\
\midrule
Onion-Greedy & 72.38 & 4.33 & 76.92 & \textbf{0.33} & Conventional \\
SimAnneal & 71.43 & 23.66 & 79.79 & 40.89 & Conventional \\
ILP-Utility & \textbf{87.05} & 14.13 & 73.70 & 32.83 & Conventional \\
GraphGreedy-Reimpl & 68.38 & 3.24 & \textbf{87.49} & 0.44 & Local recent proxy \\
DynHysteresis-Reimpl & 68.67 & \textbf{0.48} & \underline{86.91} & \underline{0.41} & Local recent proxy \\
LABELSENSE-Pilot & \underline{85.62} & \underline{2.09} & 76.47 & 27.01 & Proposed pilot \\
\bottomrule
\end{tabular}
\end{table*}

\begin{figure*}[t]
\centering
\pgfplotsset{lsaxis/.style={width=0.47\textwidth,height=4.2cm,grid=major,grid style={gray!18},tick label style={font=\scriptsize},label style={font=\scriptsize},title style={font=\scriptsize},legend style={font=\scriptsize,draw=none}}}
\begin{tikzpicture}\begin{groupplot}[group style={group size=2 by 1,horizontal sep=1.15cm},lsaxis,symbolic x coords={Onion,SA,ILP,GraphG,DynH,LS},xtick=data,x tick label style={rotate=35,anchor=east}]
\nextgroupplot[title={(a) Displayed labels},ylabel={Rate (\%)},ymin=0,ymax=100]\addplot+[ybar,fill=blue!55,draw=blue!70!black] table[x=method,y=display_pct,col sep=comma]{figures/data/fig1_main.csv};
\nextgroupplot[title={(b) Temporal flicker},ylabel={Changes (\%)},ymin=0]\addplot+[ybar,fill=orange!65,draw=orange!70!black] table[x=method,y=flicker_pct,col sep=comma]{figures/data/fig1_main.csv};
\end{groupplot}\end{tikzpicture}
\caption{Main five-seed trade-off. Rates aggregate 25 frames per method; all displayed layouts had zero counted collisions.}
\label{fig:main}
\end{figure*}
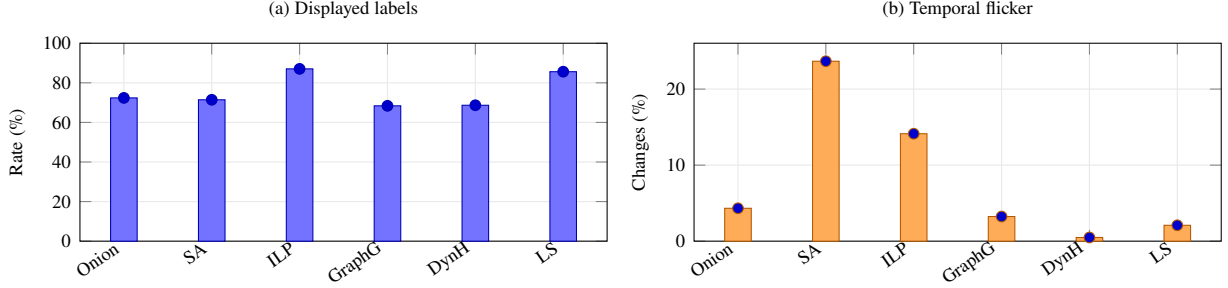

Table~\ref{tab:main-results} and Fig.~\ref{fig:main} show a mixed outcome. \method{} displays \LSDisplayPct\% of labels, with \LSFlickerPct\% flicker, \LSPrefPct\% synthetic preference agreement, and \LSRuntimeMs{} ms per frame. Every main method has a 100\% collision-free rate under the supplied rectangles and clearance. ILP-Utility displays the most labels (\ILPDisplayPct\%) but switches positions on \ILPFlickerPct\% of shared selections. Relative to that exact-handcrafted comparator, \method{} loses 1.43 display percentage points, reduces flicker by \FlickerReductionVsILP{} points (\FlickerRelativeReductionVsILP\% relative), and gains \PreferenceGainVsILP{} points on the synthetic preference metric.

The greedy learned comparators reveal the cost of constrained yield. GraphGreedy-Reimpl is approximately \RuntimeRatioVsGraph$\times$ faster and achieves higher synthetic preference agreement, but \method{} displays \DisplayGainVsGraph{} additional percentage points. DynHysteresis-Reimpl has the lowest flicker at \DynFlickerPct\%, yet displays only 68.67\% of input labels. The proposed pilot therefore occupies a high-yield, low-switching region; it is not uniformly dominant.

\begin{table*}[t]
\caption{Paired exploratory contrasts across five generated-scene seeds. Differences are percentage points (LABELSENSE-Pilot minus comparator); confidence intervals are paired bootstrap intervals. Holm-adjusted Wilcoxon values are descriptive because $n=5$.}
\label{tab:statistics}
\centering\small
\begin{tabular}{llrrr}
\toprule
Comparator & Metric & Mean diff. & 95\% bootstrap CI & Holm $p$ \\
\midrule
ILP-Utility & display rate & -1.43 & [-2.10, -0.76] & 1.000 \\
ILP-Utility & flicker rate & -12.04 & [-19.75, -5.38] & 1.000 \\
ILP-Utility & preference agreement & 2.77 & [0.83, 4.85] & 1.000 \\
GraphGreedy-Reimpl & display rate & 17.24 & [13.81, 20.00] & 1.000 \\
GraphGreedy-Reimpl & flicker rate & -1.15 & [-2.84, 0.54] & 1.000 \\
GraphGreedy-Reimpl & preference agreement & -11.02 & [-13.68, -8.83] & 1.000 \\
DynHysteresis-Reimpl & display rate & 16.95 & [13.52, 20.00] & 1.000 \\
DynHysteresis-Reimpl & flicker rate & 1.60 & [0.96, 2.27] & 1.000 \\
DynHysteresis-Reimpl & preference agreement & -10.44 & [-13.54, -7.92] & 1.000 \\
\bottomrule
\end{tabular}
\end{table*}

Exploratory paired analyses average each five-frame sequence within seed. We report paired mean differences, 20,000-sample bootstrap intervals, exact two-sided Wilcoxon tests when defined, and Holm correction across the generated contrasts. Table~\ref{tab:statistics} gives the three central comparator sets. Every adjusted value is 1.000. With $n=5$ constructed scene seeds, these values are descriptive consistency checks, not population inference or evidence of statistical equivalence.

\subsection{Motion, Text Width, Density, and Runtime}

\begin{figure*}[t]
\centering
\pgfplotsset{lsaxis/.style={width=0.47\textwidth,height=4.2cm,grid=major,grid style={gray!18},tick label style={font=\scriptsize},label style={font=\scriptsize},title style={font=\scriptsize},legend style={font=\scriptsize,draw=none}}}
\begin{tikzpicture}\begin{groupplot}[group style={group size=2 by 1,horizontal sep=1.2cm},lsaxis]
\nextgroupplot[title={(a) Pan/zoom stability},xlabel={Motion amplitude (\%)},ylabel={Consistency (\%)},ymin=80,ymax=101]\addplot+[mark=*,blue,thick] table[x=motion_pct,y=consistency_pct,col sep=comma]{figures/data/fig2_zoom.csv};
\nextgroupplot[title={(b) Label length},xlabel={Length factor},ylabel={Displayed (\%)},ymin=45,ymax=101]\addplot+[mark=square*,red!75!black,thick] table[x=length_factor,y=display_pct,col sep=comma]{figures/data/fig2_length.csv};
\end{groupplot}\end{tikzpicture}
\caption{Controlled dynamic and text-length sensitivity for \method{}. Camera motion is scripted and length factors alter generated strings.}
\label{fig:zoom-length}
\end{figure*}
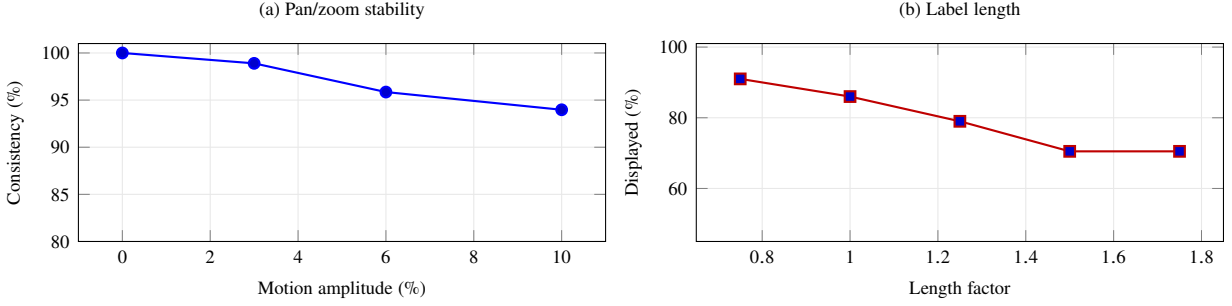

Figure~\ref{fig:zoom-length} separates camera amplitude from string length. With no motion, position consistency is 100\%; at 10\% scripted amplitude it remains 93.97\%. This behavior reflects one-step direction reuse on a deterministic camera path and cannot predict stability under real interaction traces. Increasing the length factor from 0.75 to 1.50 reduces displayed yield from 91.00\% to 70.50\%; the 1.75 condition produces the same constructed string length for this generator and therefore repeats 70.50\%.

\begin{figure*}[t]
\centering
\pgfplotsset{lsaxis/.style={width=0.47\textwidth,height=4.2cm,grid=major,grid style={gray!18},tick label style={font=\scriptsize},label style={font=\scriptsize},title style={font=\scriptsize},legend style={font=\scriptsize,draw=none}}}
\begin{tikzpicture}\begin{groupplot}[group style={group size=2 by 1,horizontal sep=1.2cm},lsaxis]
\nextgroupplot[title={(a) Density},xlabel={Features per viewport},ylabel={Displayed (\%)},ymin=40,ymax=101]\addplot+[mark=*,blue,thick] table[x=points,y=display_pct,col sep=comma]{figures/data/fig3_density.csv};
\nextgroupplot[title={(b) Runtime},xlabel={Features per viewport},ylabel={Solver time (ms)},ymode=log]\addplot+[mark=triangle*,orange!80!black,thick] table[x=points,y=runtime_ms,col sep=comma]{figures/data/fig3_density.csv};
\end{groupplot}\end{tikzpicture}
\caption{Density scalability. Runtime uses a log axis and reports the executed single-process environment, not a hardware-normalized benchmark.}
\label{fig:density}
\end{figure*}
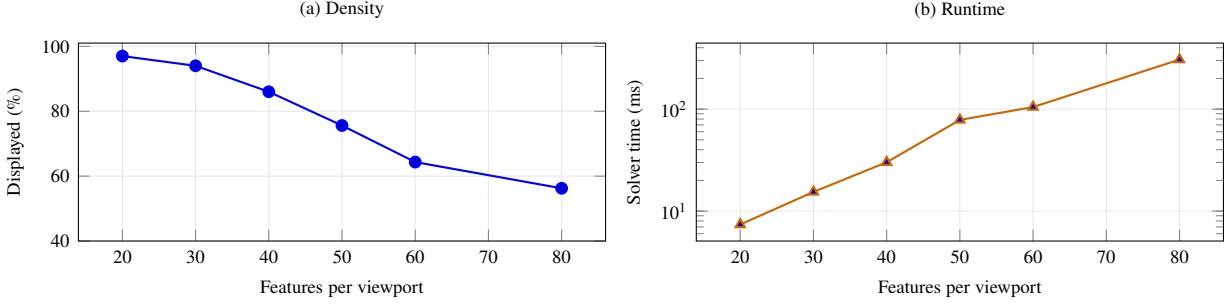

Density exposes the binary solver cost (Fig.~\ref{fig:density}). Display rate falls from 97.00\% at 20 points to \DensityEightyDisplayPct\% at 80 as rectangular conflicts increase. Mean local solver time grows from \DensityTwentyRuntimeMs{} to \DensityEightyRuntimeMs{} ms. All five 80-point runs returned HiGHS-optimal status within the configured 0.50-s limit, but the \DensityEightyRuntimeSdMs-ms standard deviation shows unstable tail behavior. Because point distance and rectangle conflicts are both built quadratically, this pilot is not a production-scale renderer.

\subsection{Preference, Suffix, and Enlarged-Geometry Stress}

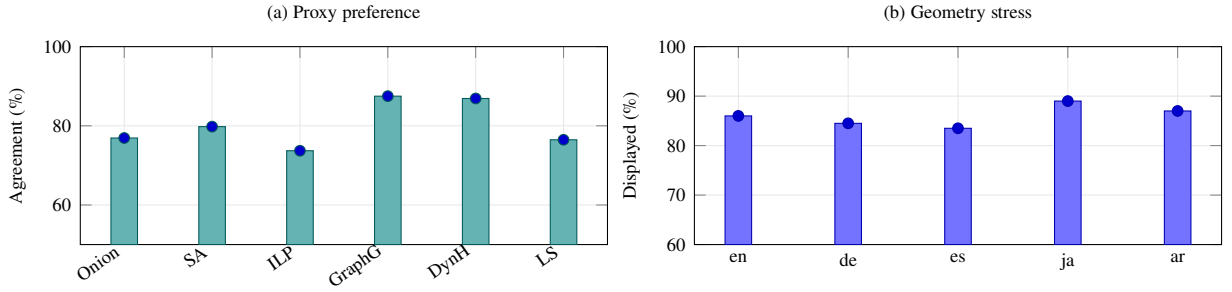
\begin{figure*}[t]
\centering
\pgfplotsset{lsaxis/.style={width=0.47\textwidth,height=4.2cm,grid=major,grid style={gray!18},tick label style={font=\scriptsize},label style={font=\scriptsize},title style={font=\scriptsize},legend style={font=\scriptsize,draw=none}}}
\begin{tikzpicture}\begin{groupplot}[group style={group size=2 by 1,horizontal sep=1.15cm},lsaxis]
\nextgroupplot[title={(a) Proxy preference},ylabel={Agreement (\%)},ymin=50,ymax=100,symbolic x coords={Onion,SA,ILP,GraphG,DynH,LS},xtick=data,x tick label style={rotate=35,anchor=east}]\addplot+[ybar,fill=teal!60,draw=teal!70!black] table[x=method,y=preference_pct,col sep=comma]{figures/data/fig1_main.csv};
\nextgroupplot[title={(b) Geometry stress},ylabel={Displayed (\%)},ymin=60,ymax=100,symbolic x coords={en,de,es,ja,ar},xtick=data]\addplot+[ybar,fill=blue!55,draw=blue!70!black] table[x=language,y=display_pct,col sep=comma]{figures/data/fig4_language.csv};
\end{groupplot}\end{tikzpicture}
\caption{Synthetic preference and script-suffix geometry stress. Panel (b) is not a multilingual usability comparison.}
\label{fig:preference-language}
\end{figure*}

Figure~\ref{fig:preference-language}(a) repeats the main synthetic preference score to make its conflict with yield visible. The learned greedy variants score highest because selection by the learned proxy can omit difficult labels. The score is a mean over selected directions, so displaying fewer labels can make it easier to achieve a high value. It must not be read as user preference.

The suffix experiment appends generic facility terms in five scripts and estimates width through Eq.~\eqref{eq:size}. Display rates range from 83.50\% for the Spanish suffix to 89.00\% for the Japanese suffix (Table~\ref{tab:robustness} and Fig.~\ref{fig:preference-language}(b)). This ordering follows constructed string widths and held-out geometry; it is not evidence that one language performs better. Source proper names were preserved, strings were not authenticated translations, and Arabic shaping, bidirectional layout, fallback fonts, truncation, and line breaking were not rendered.

\begin{table}[t]
\caption{Geometry stress results. Suffix rows are controlled strings, not authenticated translations.}
\label{tab:robustness}
\centering\scriptsize
\begin{tabular}{lrrr}
\toprule
Condition & Display & Pref. & ms \\
\midrule
Suffix-ar & 87.00 & 81.98 & 39.48 \\
Suffix-de & 84.50 & 80.49 & 26.78 \\
Suffix-en & 86.00 & 81.56 & 30.17 \\
Suffix-es & 83.50 & 78.71 & 34.56 \\
Suffix-ja & 89.00 & 82.40 & 29.58 \\
Aware 1.5$\times$ font & 79.50 & 78.51 & 31.94 \\
Unaware at 1.5$\times$ eval. & 76.50 & 90.49 & 0.41 \\
\bottomrule
\end{tabular}
\end{table}

The aware 1.5$\times$ profile constructs and optimizes enlarged rectangles, displays \AwareFifteenDisplayPct\%, and records \AwareFifteenViolationPct\% post-check violations. The unaware GraphGreedy layout is optimized at 1.0$\times$ and reevaluated at 1.5$\times$; \UnawareViolationPct\% of its selected placements violate the enlarged-box proxy. This demonstrates geometry mismatch, not accessibility or low-vision benefit. A binary violation counts a viewport overflow or conflict event per selected layout convention and is not a participant outcome.

\subsection{Ablation}

\begin{table}[t]
\caption{Component ablation (percent). Accessibility violations are enlarged-box proxy violations.}
\label{tab:ablation}
\centering\scriptsize
\begin{tabular}{lrrrr}
\toprule
Variant & Display & Flicker & Pref. & Viol. \\
\midrule
Full & 82.11 & 1.61 & 77.16 & 0.00 \\
No temporal & 82.50 & 12.93 & 77.73 & 0.00 \\
No graph context & 85.26 & 9.25 & 71.34 & 0.00 \\
No preference & 83.29 & 1.83 & 57.77 & 0.00 \\
Greedy decoder & 66.05 & 5.31 & 86.28 & 0.00 \\
Unaware geometry & 89.08 & 1.60 & 80.45 & 66.35 \\
\bottomrule
\end{tabular}
\end{table}

\begin{figure*}[t]
\centering
\pgfplotsset{lsaxis/.style={width=0.47\textwidth,height=4.2cm,grid=major,grid style={gray!18},tick label style={font=\scriptsize},label style={font=\scriptsize},title style={font=\scriptsize},legend style={font=\scriptsize,draw=none}}}
\begin{tikzpicture}\begin{groupplot}[group style={group size=2 by 1,horizontal sep=1.25cm},lsaxis]
\nextgroupplot[title={(a) Accessibility proxy},ylabel={Rate (\%)},ymin=0,ymax=100,symbolic x coords={Aware 1x,Aware 1.2x,Aware 1.5x,Unaware 1.5x},xtick=data,x tick label style={rotate=32,anchor=east},legend pos=north west]\addplot+[ybar,bar width=4pt,fill=blue!55] table[x=profile,y=display_pct,col sep=comma]{figures/data/fig5_accessibility.csv};\addlegendentry{Displayed}\addplot+[ybar,bar width=4pt,fill=red!60] table[x=profile,y=violations_pct,col sep=comma]{figures/data/fig5_accessibility.csv};\addlegendentry{Violations}
\nextgroupplot[title={(b) Ablation},xlabel={Flicker (\%)},ylabel={Displayed (\%)},ymin=60,ymax=100]\addplot+[only marks,mark=*,blue] table[x=flicker_pct,y=display_pct,col sep=comma]{figures/data/fig5_ablation.csv};
\end{groupplot}\end{tikzpicture}
\caption{Accessibility-inspired geometry stress and component ablation. ``Aware'' means the solver used the evaluated box size; no participant accessibility outcome was collected.}
\label{fig:access-ablation}
\end{figure*}
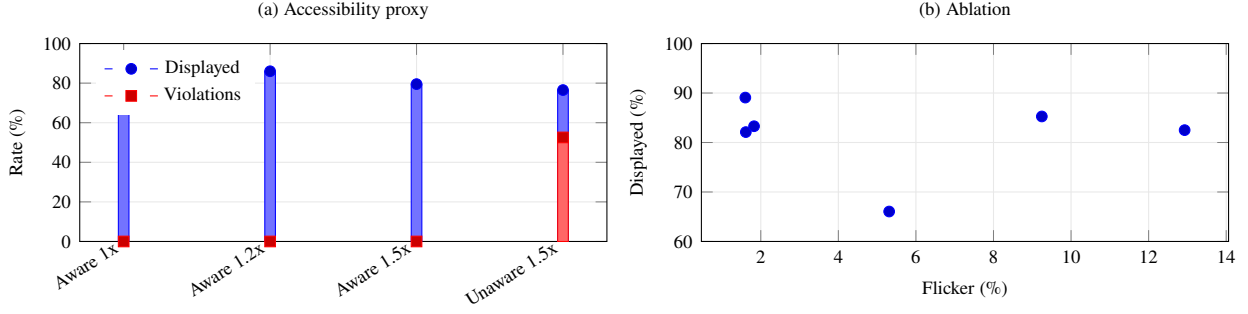

Removing temporal utility raises flicker from \FullAblationFlickerPct\% to \NoTemporalFlickerPct\%, a \TemporalFlickerReduction-point reduction for the full configuration, while display rate changes from 82.50\% to 82.11\%. Removing the preference term drops synthetic agreement by 19.39 points. Replacing learned graph-context scoring with handcrafted ILP raises display but also flicker and reduces the proxy score. The unaware-geometry variant displays 89.08\% at its smaller construction geometry, then incurs \UnawareAblationViolationPct\% violations at 1.5$\times$ evaluation. Figure~\ref{fig:access-ablation} makes the yield--validity and yield--switching trade-offs explicit.

H1--H3 pass only in their narrow engineering form: the proposed decoder displays more labels than learned greedy with audited geometric feasibility; its temporal term substantially reduces switching; and enlargement-aware construction avoids the violations induced by mismatched boxes. These outcomes do not validate the original graph-transformer or human-centered hypotheses.

\section{Discussion}
\label{sec:discussion}

\subsection{What the Pilot Establishes}

The results support one systems proposition: when every method shares a candidate set, a learned utility can be separated cleanly from a hard-feasibility decoder. The MILP never needs to trust the scorer about geometry; invalid candidates receive zero upper bounds and every conflict pair is constrained. This separation permits preference and stability weights to change without weakening the encoded rectangle guarantee. It also makes failure legible: a layout can be geometrically valid while scoring poorly, switching often, hiding too many labels, or failing a later enlarged-box audit.

The comparison with ILP-Utility is especially informative. The proposed scoring sacrifices only 1.43 display points while cutting switching sharply, but that result is partly designed into the one-step bonus. It demonstrates that the implementation obeys its objective, not that people prefer the resulting sequence. The even lower flicker of DynHysteresis-Reimpl shows that a greedy persistence bias can be effective when lower display is acceptable. A future interface should expose a Pareto frontier instead of claiming one universally best weight.

The learned component also has a restricted interpretation. The MLP predicts a synthetic target built from its own inputs, so low training RMSE confirms implementation consistency rather than discovery of latent cartographic taste. The graph-context terms are fixed radial averages. They summarize competition but do not pass candidate-level messages, learn attention weights, or model multiple layers of interaction. Calling this implementation a graph transformer would therefore be inaccurate.

\subsection{Threats to Validity}

\textit{Construct validity.} Display and collision metrics concern axis-aligned heuristic boxes. They omit glyph outlines, halos, markers, leaders, color contrast, occlusion by map content, screen-reader structure, and point--label ambiguity. Synthetic preference agreement is conditional on displayed labels and may reward selective omission. Flicker counts compass-direction changes only among objects visible in consecutive frames; appearance, disappearance, and residual motion are not penalized.

\textit{Data validity.} Airports differ materially from urban points of interest in density, naming, priority, and spatial pattern. The pinned prefix is not a random global sample. Dense scenes depend on replication and jitter. Whole-country splitting limits direct leakage but five held-out countries do not establish cross-region generalization. Generic facility suffixes neither validate translation nor represent real localized map strings.

\textit{Human and accessibility validity.} No participant was recruited, no human response file was analyzed, and no task-completion time, comprehension, error, readability, or preference outcome was collected. Enlarged-font rectangles are an accessibility-inspired geometry proxy only. Passing the proxy does not establish WCAG conformance or accessible map use \cite{manu2025accessibility,madugalla2025adaptive}. The identified preference dataset contains valuable prior responses, but it was not acquired in this run and would not substitute for multilingual and low-vision validation \cite{scheuerman2023dataset}.

\textit{Comparator validity.} Onion-Greedy, SimAnneal, and ILP-Utility are local implementations. GraphGreedy-Reimpl and DynHysteresis-Reimpl capture broad scoring and hysteresis ideas but are not official recent systems. No official LPGT, reinforced-label, Rapid Labels, or temporal-labeling code was executed. Candidate sets, objectives, and hardware may differ from the cited literature, so the table supports only internal matched comparisons.

\textit{Inference and performance validity.} Five generated-scene seeds yield low-powered paired summaries; Holm-adjusted $p$ values of 1.000 preclude significance claims. Bootstrap intervals describe these seeds, not a population. Runtime was measured on one shared virtualized CPU with no renderer or network pipeline. Means are not service-level p95/p99 guarantees.

\section{Conclusion}
\label{sec:conclusion}

LABELSENSE-Pilot demonstrates an integration of graph-context utility scoring, one-frame temporal persistence, profile-dependent label rectangles, and hard-constrained point-feature selection. On five generated held-out-country sequences, the method displayed 85.62 percent of labels, changed 2.09 percent of shared placements, and produced no collisions under encoded boxes. Compared with handcrafted MILP utility, it traded 1.43 display percentage points for a 12.04-point flicker reduction. Constructing labels at 1.5 times font geometry avoided proxy violations, whereas layouts built at standard geometry and reevaluated after enlargement violated 52.57 percent of selected placements. These findings verify software behavior and expose yield, stability, preference-proxy, and runtime trade-offs. They do not demonstrate accessible map use, multilingual usability, human preference, comprehension, or task performance. The executed learner is an MLP over fixed graph-context summaries, not a graph transformer; language strings, preference targets, density, and motion are generated stressors; and no official recent implementation was executed. A publication-ready study must add independent urban regions, authentic localized strings with pinned shaping, the intended graph model, official baselines, observed interaction traces, and preregistered participation by multilingual and low-vision users. Until those requirements are met and venue policies are rechecked, the Q1 submission decision remains NO-GO.

\balance
\bibliographystyle{IEEEtran}
\bibliography{references}

@article{zhang2026lpce,
  author  = {Zhang, Pingshun and Che, Enyu and Chen, Yinan and Huang, Bingyao and Ling, Haibin and Qu, Jingwei},
  title   = {Mixture of Cluster-guided Experts for Retrieval-Augmented Label Placement},
  journal = {IEEE Transactions on Visualization and Computer Graphics},
  year    = {2026},
  volume  = {32},
  number  = {1},
  pages   = {418--428},
  doi     = {10.1109/TVCG.2025.3642518},
  url     = {https://ieeexplore.ieee.org/document/11295938}
}

@article{qu2025graph,
  author  = {Qu, Jingwei and Zhang, Pingshun and Che, Enyu and Chen, Yinan and Ling, Haibin},
  title   = {Graph Transformer for Label Placement},
  journal = {IEEE Transactions on Visualization and Computer Graphics},
  year    = {2025},
  volume  = {31},
  number  = {1},
  pages   = {1257--1267},
  doi     = {10.1109/TVCG.2024.3456141},
  url     = {https://ieeexplore.ieee.org/document/10670468},
  note    = {Presented at IEEE VIS 2024; the application domain is household-appliance manual illustrations, not maps}
}

@article{bobak2024reinforced,
  author  = {Bob{\'a}k, Petr and \v{C}mol{\'i}k, Ladislav and \v{C}ad{\'i}k, Martin},
  title   = {Reinforced Labels: Multi-Agent Deep Reinforcement Learning for Point-Feature Label Placement},
  journal = {IEEE Transactions on Visualization and Computer Graphics},
  year    = {2024},
  volume  = {30},
  number  = {9},
  pages   = {5908--5922},
  doi     = {10.1109/TVCG.2023.3313729},
  url     = {https://ieeexplore.ieee.org/document/10246431}
}

@article{chen2023rllabel,
  author  = {Chen, Zhu-Tian and Chiappalupi, Daniele and Lin, Tica and Yang, Yalong and Beyer, Johanna and Pfister, Hanspeter},
  title   = {{RL-LABEL}: A Deep Reinforcement Learning Approach Intended for {AR} Label Placement in Dynamic Scenarios},
  journal = {IEEE Transactions on Visualization and Computer Graphics},
  year    = {2023},
  doi     = {10.1109/TVCG.2023.3326568},
  url     = {https://ieeexplore.ieee.org/document/10290983},
  note    = {Online-ahead-of-print record; presented at IEEE VIS 2023. Verify final issue metadata before submission}
}

@article{pavlovec2022rapid,
  author  = {Pavlovec, V{\'a}clav and \v{C}mol{\'i}k, Ladislav},
  title   = {Rapid Labels: Point-Feature Labeling on {GPU}},
  journal = {IEEE Transactions on Visualization and Computer Graphics},
  year    = {2022},
  volume  = {28},
  number  = {1},
  pages   = {604--613},
  doi     = {10.1109/TVCG.2021.3114854},
  url     = {https://ieeexplore.ieee.org/document/9552249}
}

@article{haunert2017beyond,
  author  = {Haunert, Jan-Henrik and Wolff, Alexander},
  title   = {Beyond Maximum Independent Set: An Extended Integer Programming Formulation for Point Labeling},
  journal = {ISPRS International Journal of Geo-Information},
  year    = {2017},
  volume  = {6},
  number  = {11},
  pages   = {342},
  doi     = {10.3390/ijgi6110342},
  url     = {https://www.mdpi.com/2220-9964/6/11/342}
}

@article{cao2023spatial,
  author  = {Cao, Wen and Xu, Jiaqi and Peng, Feilin and Tong, Xiaochong and Wang, Xinyi and Zhao, Siqi and Liu, Wenhao},
  title   = {A Point-Feature Label Placement Algorithm Based on Spatial Data Mining},
  journal = {Mathematical Biosciences and Engineering},
  year    = {2023},
  volume  = {20},
  number  = {7},
  pages   = {12169--12193},
  doi     = {10.3934/mbe.2023542},
  url     = {https://www.aimspress.com/article/doi/10.3934/mbe.2023542}
}

@article{lessani2021parallel,
  author  = {Lessani, Mohammad Naser and Deng, Jiqiu and Guo, Zhiyong},
  title   = {A Novel Parallel Algorithm with Map Segmentation for Multiple Geographical Feature Label Placement Problem},
  journal = {ISPRS International Journal of Geo-Information},
  year    = {2021},
  volume  = {10},
  number  = {12},
  pages   = {826},
  doi     = {10.3390/ijgi10120826},
  url     = {https://www.mdpi.com/2220-9964/10/12/826}
}

@article{lessani2025mpi,
  author  = {Lessani, M. Naser and Li, Zhenlong and Deng, Jiqiu and Guo, Zhiyong},
  title   = {An {MPI}-Based Parallel Genetic Algorithm for Multiple Geographical Feature Label Placement Based on the Hybrid of Fixed-Sliding Models},
  journal = {Geo-spatial Information Science},
  year    = {2025},
  volume  = {28},
  number  = {2},
  pages   = {761--779},
  doi     = {10.1080/10095020.2024.2313326},
  url     = {https://doi.org/10.1080/10095020.2024.2313326},
  note    = {Published online in 2024}
}

@article{zoraster1990solution,
  author  = {Zoraster, Steven},
  title   = {The Solution of Large 0--1 Integer Programming Problems Encountered in Automated Cartography},
  journal = {Operations Research},
  year    = {1990},
  volume  = {38},
  number  = {5},
  pages   = {752--759},
  doi     = {10.1287/opre.38.5.752},
  url     = {https://doi.org/10.1287/opre.38.5.752}
}

@article{klau2003optimal,
  author  = {Klau, Gunnar W. and Mutzel, Petra},
  title   = {Optimal Labeling of Point Features in Rectangular Labeling Models},
  journal = {Mathematical Programming},
  year    = {2003},
  volume  = {94},
  pages   = {435--458},
  doi     = {10.1007/s10107-002-0327-9},
  url     = {https://doi.org/10.1007/s10107-002-0327-9}
}

@article{bonerath2025timeslider,
  author  = {Bonerath, Annika and Driemel, Anne and Haunert, Jan-Henrik and Haverkort, Herman and Langetepe, Elmar and Niedermann, Benjamin},
  title   = {Algorithms for Consistent Dynamic Labeling of Maps With a Time-Slider Interface},
  journal = {IEEE Transactions on Visualization and Computer Graphics},
  year    = {2025},
  volume  = {31},
  number  = {10},
  pages   = {6691--6704},
  doi     = {10.1109/TVCG.2025.3527582},
  url     = {https://ieeexplore.ieee.org/document/10834546}
}

@inproceedings{cutello2025tabu,
  author    = {Cutello, Vincenzo and Mezzina, A. and Pavone, Mario and Zito, Francesco},
  title     = {A Real-Time Adaptive Tabu Search for Handling Zoom In/Out in Map Labeling Problem},
  booktitle = {Learning and Intelligent Optimization: 18th International Conference, LION 2024},
  series    = {Lecture Notes in Computer Science},
  volume    = {14990},
  pages     = {108--122},
  publisher = {Springer},
  year      = {2025},
  doi       = {10.1007/978-3-031-75623-8_9},
  url       = {https://link.springer.com/chapter/10.1007/978-3-031-75623-8_9}
}

@inproceedings{depian2023transitions,
  author    = {Depian, Thomas and Li, Guangping and N{\"o}llenburg, Martin and Wulms, Jules},
  title     = {Transitions in Dynamic Point Labeling},
  booktitle = {12th International Conference on Geographic Information Science (GIScience 2023)},
  series    = {Leibniz International Proceedings in Informatics (LIPIcs)},
  volume    = {277},
  pages     = {2:1--2:19},
  publisher = {Schloss Dagstuhl -- Leibniz-Zentrum f{\"u}r Informatik},
  year      = {2023},
  doi       = {10.4230/LIPIcs.GIScience.2023.2},
  url       = {https://drops.dagstuhl.de/entities/document/10.4230/LIPIcs.GIScience.2023.2}
}

@article{he2022smoothness,
  author  = {He, Y. and Zhao, G.-D. and Zhang, S.-H.},
  title   = {Smoothness Preserving Layout for Dynamic Labels by Hybrid Optimization},
  journal = {Computational Visual Media},
  year    = {2022},
  volume  = {8},
  pages   = {149--163},
  doi     = {10.1007/s41095-021-0231-y},
  url     = {https://link.springer.com/article/10.1007/s41095-021-0231-y}
}

@article{bhore2022dynamic,
  author  = {Bhore, Sujoy and Li, Guangping and N{\"o}llenburg, Martin},
  title   = {An Algorithmic Study of Fully Dynamic Independent Sets for Map Labeling},
  journal = {ACM Journal of Experimental Algorithmics},
  year    = {2022},
  volume  = {27},
  pages   = {1.8:1--1.8:36},
  doi     = {10.1145/3514240},
  url     = {https://dl.acm.org/doi/10.1145/3514240}
}

@article{gedicke2021zoomless,
  author  = {Gedicke, Sven and Bonerath, Annika and Niedermann, Benjamin and Haunert, Jan-Henrik},
  title   = {Zoomless Maps: External Labeling Methods for the Interactive Exploration of Dense Point Sets at a Fixed Map Scale},
  journal = {IEEE Transactions on Visualization and Computer Graphics},
  year    = {2021},
  volume  = {27},
  number  = {2},
  doi     = {10.1109/TVCG.2020.3030379},
  url     = {https://ieeexplore.ieee.org/document/9222088}
}

@article{gemsa2020unified,
  author  = {Gemsa, Andreas and Niedermann, Benjamin and N{\"o}llenburg, Martin},
  title   = {A Unified Model and Algorithms for Temporal Map Labeling},
  journal = {Algorithmica},
  year    = {2020},
  volume  = {82},
  pages   = {2709--2736},
  doi     = {10.1007/s00453-020-00694-7},
  url     = {https://link.springer.com/article/10.1007/s00453-020-00694-7}
}

@inproceedings{barth2016temporal,
  author    = {Barth, Lukas and Niedermann, Benjamin and N{\"o}llenburg, Martin and Strash, Darren},
  title     = {Temporal Map Labeling: A New Unified Framework with Experiments},
  booktitle = {Proceedings of the 24th ACM SIGSPATIAL International Conference on Advances in Geographic Information Systems},
  pages     = {23:1--23:10},
  publisher = {ACM},
  year      = {2016},
  doi       = {10.1145/2996913.2996957},
  url       = {https://dl.acm.org/doi/10.1145/2996913.2996957}
}

@article{been2006dynamic,
  author  = {Been, Ken and Daiches, Eli and Yap, Chee-Keng},
  title   = {Dynamic Map Labeling},
  journal = {IEEE Transactions on Visualization and Computer Graphics},
  year    = {2006},
  volume  = {12},
  number  = {5},
  pages   = {773--780},
  doi     = {10.1109/TVCG.2006.136},
  url     = {https://doi.org/10.1109/TVCG.2006.136}
}

@article{scheuerman2023dataset,
  author  = {Scheuerman, Jaelle and Harman, Jason L. and Goldstein, Rebecca R. and Acklin, Dina and Michael, Chris J.},
  title   = {Label Placement Preferences for Digital Maps},
  journal = {Discover Psychology},
  year    = {2023},
  volume  = {3},
  pages   = {23},
  doi     = {10.1007/s44202-023-00081-7},
  url     = {https://link.springer.com/article/10.1007/s44202-023-00081-7}
}

@article{scheuerman2023visual,
  author  = {Scheuerman, Jaelle and Harman, Jason L. and Goldstein, Rebecca R. and Acklin, Dina and Michael, Chris J.},
  title   = {Visual Preferences in Map Label Placement},
  journal = {Discover Psychology},
  year    = {2023},
  volume  = {3},
  pages   = {27},
  doi     = {10.1007/s44202-023-00088-0},
  url     = {https://link.springer.com/article/10.1007/s44202-023-00088-0}
}

@article{harrie2022challenges,
  author  = {Harrie, Lars and Oucheikh, Rachid and Nilsson, {\AA}sa and Oxenstierna, Andreas and Cederholm, Pontus and Wei, Lai and Richter, Kai-Florian and Olsson, Perola},
  title   = {Label Placement Challenges in City Wayfinding Map Production---Identification and Possible Solutions},
  journal = {Journal of Geovisualization and Spatial Analysis},
  year    = {2022},
  volume  = {6},
  pages   = {16},
  doi     = {10.1007/s41651-022-00115-z},
  url     = {https://link.springer.com/article/10.1007/s41651-022-00115-z}
}

@article{madugalla2025adaptive,
  author  = {Madugalla, Anuradha and Huang, Yutan and Grundy, John and Cho, Min Hee and Gamage, Lasith Koswatta and Lau, Y. P. and Leao, Tristan and Thiele, Sam},
  title   = {Engineering Support for Adaptations in Information Graphics for Disabled Communities: A Study with Public Space Indoor Maps},
  journal = {Empirical Software Engineering},
  year    = {2025},
  volume  = {30},
  pages   = {160},
  doi     = {10.1007/s10664-025-10715-0},
  url     = {https://link.springer.com/article/10.1007/s10664-025-10715-0}
}

@article{manu2025accessibility,
  author  = {Manu, Samuel D. and Burghardt, Dirk and Hauthal, Eva},
  title   = {Enhancing Accessibility of Thematic Web Maps for Visually Impaired Users},
  journal = {KN - Journal of Cartography and Geographic Information},
  year    = {2025},
  volume  = {75},
  pages   = {107--121},
  doi     = {10.1007/s42489-025-00189-x},
  url     = {https://link.springer.com/article/10.1007/s42489-025-00189-x}
}

@inproceedings{li2024altgeoviz,
  author    = {Li, Chu and Pang, Rock Yuren and Sharif, Ather and Chheda-Kothary, Arnavi and Heer, Jeffrey and Froehlich, Jon E.},
  title     = {{AltGeoViz}: Facilitating Accessible Geovisualization},
  booktitle = {2024 IEEE Visualization and Visual Analytics (VIS)},
  pages     = {61--65},
  publisher = {IEEE},
  year      = {2024},
  doi       = {10.1109/VIS55277.2024.00020},
  url       = {https://doi.org/10.1109/VIS55277.2024.00020}
}

@inproceedings{kulyukin2010svg,
  author    = {Kulyukin, Vladimir and Marston, James R. and Miele, Joshua A. and Kutiyanawala, Ammar},
  title     = {Automated {SVG} Map Labeling for Customizable Large Print Maps for Low Vision Individuals},
  booktitle = {Proceedings of the RESNA 2010 Annual Conference},
  year      = {2010},
  url       = {https://www.resna.org/sites/default/files/legacy/conference/proceedings/2010/TechnologyCognitiveSensory/KulyukinV2.html},
  note      = {No DOI assigned}
}

@article{hennig2017accessible,
  author  = {Hennig, Sabine and Zobl, Franz and Wasserburger, Wolfgang W.},
  title   = {Accessible Web Maps for Visually Impaired Users: Recommendations and Example Solutions},
  journal = {Cartographic Perspectives},
  year    = {2017},
  number  = {88},
  pages   = {6--27},
  doi     = {10.14714/CP88.1391},
  url     = {https://cartographicperspectives.org/index.php/journal/article/view/cp88-hennig-et-al}
}

@article{been2010active,
  author  = {Been, Ken and N{\"o}llenburg, Martin and Poon, Sheung-Hung and Wolff, Alexander},
  title   = {Optimizing Active Ranges for Consistent Dynamic Map Labeling},
  journal = {Computational Geometry},
  year    = {2010},
  volume  = {43},
  number  = {3},
  pages   = {312--328},
  doi     = {10.1016/j.comgeo.2009.03.006},
  url     = {https://doi.org/10.1016/j.comgeo.2009.03.006}
}

@article{christensen1995empirical,
  author  = {Christensen, Jon and Marks, Joe and Shieber, Stuart},
  title   = {An Empirical Study of Algorithms for Point-Feature Label Placement},
  journal = {ACM Transactions on Graphics},
  year    = {1995},
  volume  = {14},
  number  = {3},
  pages   = {203--232},
  doi     = {10.1145/212332.212334},
  url     = {https://dl.acm.org/doi/10.1145/212332.212334}
}

@article{a5,
  title={Web data retrieval: solving spatial range queries using k-nearest neighbor searches},
  author={Bae, Wan D and Alkobaisi, Shayma and Kim, Seon Ho and Narayanappa, Sada and Shahabi, Cyrus},
  journal={GeoInformatica},
  volume={13},
  number={4},
  pages={483--514},
  year={2009},
  publisher={Springer}
}

@article{a6,
  title={Predicting health risks of adult asthmatics susceptible to indoor air quality using improved logistic and quantile regression models},
  author={Bae, Wan D and Alkobaisi, Shayma and Horak, Matthew and Park, Choon-Sik and Kim, Sungroul and Davidson, Joel},
  journal={Life},
  volume={12},
  number={10},
  pages={1631},
  year={2022},
  publisher={MDPI}
}

@inproceedings{a8,
  title={The Tornado model: Uncertainty model for continuously changing data},
  author={Yu, Byunggu and Kim, Seon Ho and Alkobaisi, Shayma and Bae, Wan D and Bailey, Thomas},
  booktitle={International Conference on Database Systems for Advanced Applications},
  pages={624--636},
  year={2007},
  organization={Springer}
}

@article{a9,
  title={Open source computer game application: An empirical analysis of quality concerns},
  author={Ahmed, Faheem and Zia, Muhammad and Mahmood, Hasan and Al Kobaisi, Shayma},
  journal={Entertainment Computing},
  volume={21},
  pages={1--10},
  year={2017},
  publisher={Elsevier}
}

@article{a10,
  title={An interactive framework for spatial joins: a statistical approach to data analysis in GIS},
  author={Alkobaisi, Shayma and Bae, Wan D and Vojt{\v{e}}chovsk{\`y}, Petr and Narayanappa, Sada},
  journal={GeoInformatica},
  volume={16},
  number={2},
  pages={329--355},
  year={2012},
  publisher={Springer}
}

@inproceedings{a11,
  title={An interactive framework for raster data spatial joins},
  author={Bae, Wan D and Vojt{\v{e}}chovsk{\`y}, Petr and Alkobaisi, Shayma and Leutenegger, Scott T and Kim, Seon Ho},
  booktitle={Proceedings of the 15th annual ACM international symposium on Advances in geographic information systems},
  pages={1--8},
  year={2007}
}

@article{a12,
  title={Artificial intelligence algorithms in asthma management: a review of data engineering, predictive models, and future implications},
  author={Alkobaisi, Shayma and Safdar, Muhammad Farhan and Pa{\l}ka, Piotr and Abu Ali, Najah Abed},
  journal={Applied Sciences},
  volume={15},
  number={7},
  pages={3609},
  year={2025},
  publisher={MDPI}
}

\end{document}